\documentclass[9pt, conference, comsoc]{IEEEtran}
\IEEEoverridecommandlockouts
\usepackage{cite}
\usepackage{amsmath,amssymb,amsfonts}
\usepackage{algorithmic}
\usepackage{graphicx}
\usepackage{textcomp}
\usepackage{xcolor}
\usepackage{hyperref}
\usepackage{booktabs}
\usepackage{makecell}
\usepackage{fontawesome}
\usepackage[singlelinecheck=false]{caption}
\usepackage[capitalize]{cleveref}
\crefname{section}{Sec.}{Secs.}
\Crefname{section}{Section}{Sections}
\crefname{figure}{Fig.}{Figs.}
\Crefname{figure}{Figure}{Figures}

\def\BibTeX{{\rm B\kern-.05em{\sc i\kern-.025em b}\kern-.08em
    T\kern-.1667em\lower.7ex\hbox{E}\kern-.125emX}}

\begin{document}

\title{Temporally Aligned Audio for Video with Autoregression
}

\author{\IEEEauthorblockN{1\textsuperscript{st} Ilpo Viertola}
\IEEEauthorblockA{\textit{Computing Sciences} \\
\textit{Tampere University}\\
Tampere, Finland \\
ilpo.viertola@tuni.fi}
\and
\IEEEauthorblockN{2\textsuperscript{nd} Vladimir Iashin}
\IEEEauthorblockA{\textit{Computing Sciences, Tampere University} \\
\textit{Engineering Science, University of Oxford} \\
Oxford, United Kingdom \\
vi@robots.ox.ac.uk}
\and
\IEEEauthorblockN{3\textsuperscript{rd} Esa Rahtu}
\IEEEauthorblockA{\textit{Computing Sciences} \\
\textit{Tampere University}\\
Tampere, Finland \\
esa.rahtu@tuni.fi}
}
\maketitle
\begin{abstract}
We introduce V-AURA, the first autoregressive model to achieve high temporal alignment and relevance in video-to-audio generation.
V-AURA uses a high-framerate visual feature extractor and a cross-modal audio-visual feature fusion strategy to capture fine-grained visual motion events and ensure precise temporal alignment.
Additionally, we propose VisualSound, a benchmark dataset with high audio-visual relevance.
VisualSound is based on VGGSound, a video dataset consisting of in-the-wild samples extracted from YouTube.
During the curation, we remove samples where auditory events are not aligned with the visual ones.
V-AURA outperforms current state-of-the-art models in temporal alignment and semantic relevance while maintaining comparable audio quality. 
Code, samples, VisualSound and models are available at \textbf{\color{blue}\url{v-aura.notion.site}}.
\end{abstract}


\section{Introduction}
\label{sec:intro}
Video-to-audio generation focuses on synthesizing audio based on a video sequence.
The synthesized audio must be high-quality and closely aligned with the visual events temporally and semantically.
It requires the generative model to deeply understand the timing and meaning of the visual content, supported by well-curated training data where the auditory events are relevant to the visual ones.

The current state-of-the-art models are built on top of diffusion and rectified flow matching (RFM) based methods \cite{wang2023v2amapperlightweightsolutionvisiontoaudio, luo2023difffoley, wang2024frierenefficientvideotoaudiogeneration, zhang2024foleycrafterbringsilentvideos, xing2024seeinghearingopendomainvisualaudio} and have replaced autoregressive methods \cite{SpecVQGAN_Iashin_2021, mei2023foleygenvisuallyguidedaudiogeneration, sheffer2023iheartruecolors, du2023conditionalgenerationaudiovideo} built on top of Transformer architecture \cite{vaswani2023attentionneed}.
However, diffusion models often require added complexity compared to autoregressive solutions.
In particular, diffusion models rely on image-based approaches in audio generation, encoding the audio as a mel-spectrogram.
Converting audio into a mel-spectrogram is a lossy conversion as the original signal must be filtered. 
Converting the mel-spectrogram back to a waveform requires an additional network since the discarded frequencies must be reconstructed.
During the conversion, some important fine-grained audio information might be lost and generation of detailed audio becomes impossible.
In contrast, we use a pretrained audio codec that encodes waveforms into discrete token sequences and decodes them into waveform representations without transforming the audio to image-space (spectrogram) \cite{kumar2023highfidelityaudiocompressionimproved}.
Additionally, the training of diffusion models is more complex and requires more iterations than our autoregressive method.

Most of the video-to-audio models are trained with video datasets consisting of noisy in-the-wild samples scraped from YouTube, such as VGGSound \cite{chen2020vggsound} or AudioSet \cite{7952261}, where the audio-visual relevance is not guaranteed.
For example, the original audio can be replaced with non-related ones, such as background music, narration, or audio effects.
We introduce a novel benchmark, VisualSound, a subset of VGGSound in which samples possess a high audio-visual correspondence.
Removing the harmful samples, and training the model with a smaller but high-quality dataset, increases relevance and temporal alignment between video and audio.
Also, as the amount of noisy samples scales down, the training time decreases significantly.

Compared to the state-of-the-art diffusion and RFM-based models, our autoregressive approach achieves a high temporal alignment and relevance between audio and video.
Our contributions can be summarized as follows: i) the first autoregressive model to achieve strong relevance and temporal alignment in video-to-audio generation, ii) a cross-modal feature alignment strategy emphasizing the natural co-occurrence of audio and video in the autoregressive setting, iii) a new benchmark dataset with strong audio-visual correspondence, and iv) a new synchronization-based objective metric for temporal alignment between video and generated audio.

\section{Related Work}
\label{sec:rel_work}
Early approaches in visual-to-audio generation used Generative Adversarial Networks (GAN) to generate audio within a small set of data classes, given a conditional visual feature sequence \cite{ghose2021foleyganvisuallyguidedgenerative, Chen_2020}.
\cite{SpecVQGAN_Iashin_2021, mei2023foleygenvisuallyguidedaudiogeneration, sheffer2023iheartruecolors, du2023conditionalgenerationaudiovideo} framed the conditional audio generation as a next token prediction problem using various visual features as the conditional prompt.
Even though these autoregressive methods supported a wider range of audio data classes, they suffered from poor temporal alignment and audio quality.
To improve sample quality, others have explored bridging large pretrained general-purpose generative audio models to multiple modalities via feature mapping \cite{wang2023v2amapperlightweightsolutionvisiontoaudio} or by training diffusion latent aligners for semantical and temporal control \cite{xing2024seeinghearingopendomainvisualaudio, zhang2024foleycrafterbringsilentvideos}.
To emphasize temporal alignment, recent diffusion \cite{luo2023difffoley} and rectified flow matching \cite{wang2024frierenefficientvideotoaudiogeneration} approaches rely on contrastively trained audio-visual feature extractors.
In addition, \cite{wang2024frierenefficientvideotoaudiogeneration} introduces cross-modal feature fusion in an RFM setting to emphasize temporal alignment.

Even so, existing methods often fail to produce temporally aligned audio due to low visual framerates or weak learned relationships between the modalities.
To address this, we use videos with a 6-times higher framerate than the state-of-the-art and a high-framerate visual feature extractor designed to focus on fine-grain visual and motion features associated with sounds.
We enforce natural audio-video co-occurrence by introducing a channel-wise cross-modal feature fusion in an autoregressive setting.
To enhance sample quality and mitigate hallucinations caused by noisy training data, we introduce VisualSound, a novel dataset with strong audio-visual relevance.

\begin{figure}[ht]
    \centering
    \includegraphics[width=0.45\textwidth]{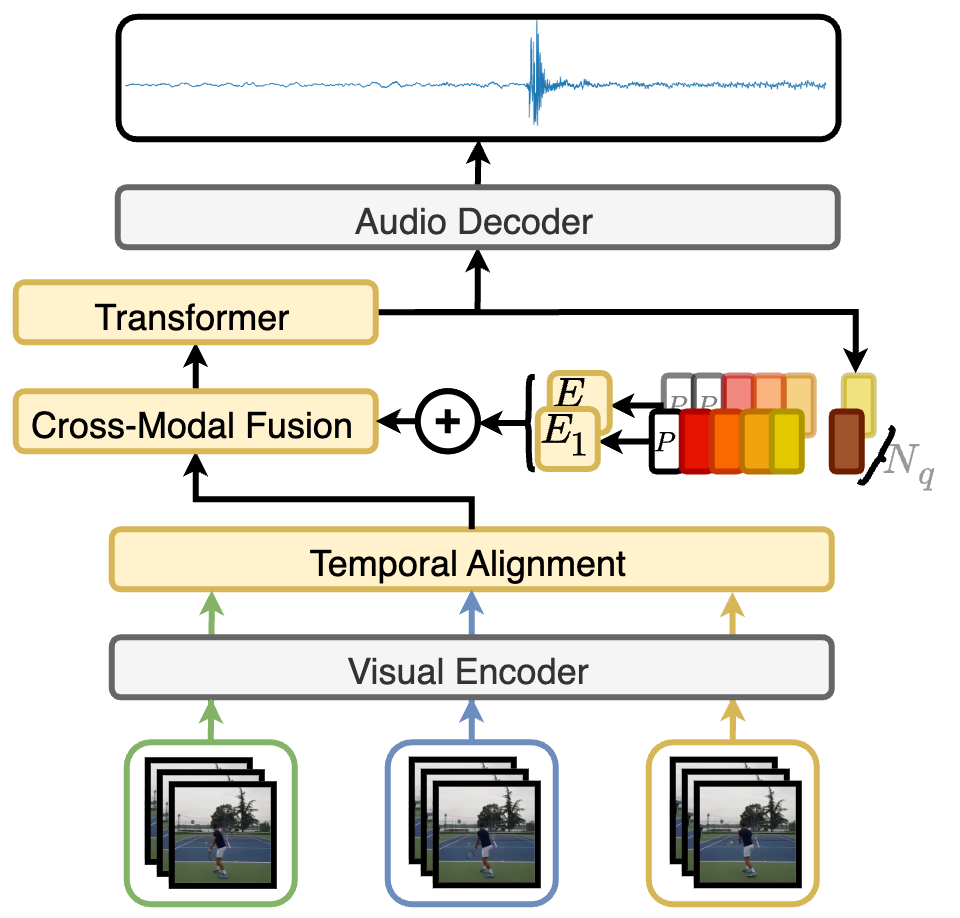}
    \vspace{-1ex}
    \caption{
        \textbf{Overview of V-AURA}. 
        Given stacks of RGB frames, the visual encoder extracts visual features which are projected into visual feature embeddings.
        Then, the temporal dimension of visual embeddings is aligned with the audio embeddings.
        The audio tokens from the previous generation step are embedded and added together to represent the full-band audio signal \cite{kumar2023highfidelityaudiocompressionimproved}.
        The tokenized audio sequence is padded with learned padding tokens ($P$).
        Embeddings of different modalities are aligned and fused with cross-modal feature fusion before the next generation step in Transformer.
        When the audio sequence reaches the desired length, it is decoded back to a waveform using the decoder of the pre-trained codebook-based autoencoder.
    }
    \vspace{-2ex}
    \label{fig:model}
\end{figure}

\section{Method}
\label{sec:method}
The proposed model, V-AURA (\textbf{V}ideo-to-\textbf{A}udio A\textbf{u}to\textbf{r}egressive Fr\textbf{a}mework), generates semantically and temporally aligned audio given a visual stream by predicting audio tokens encoded with a high-fidelity neural audio codec.
First, we extract fine-grained visual and motion features from an input video and temporally upsample them to match the temporal dimension of the audio.
Then, the temporally aligned audio and visual features are fused to emphasize the natural co-occurrence of audio and video.
Given the cross-modal feature embedding, our autoregressive model predicts the next audio token. Once the sequence reaches the desired length, it is decoded into a waveform representation.
We train our model on the VisualSound dataset, which is curated to ensure strong audio-visual relevance.
The dataset is introduced in \cref{sec:visualsound}.

\subsection{Visual Encoder}
\label{ssec:v_feat_ex}
Video-to-audio generation requires not only global but also fine-grained visual and motion information.
It is not enough to detect playing tennis in the scene; the action of the racket hitting the ball must also be caught to generate relevant audio.
To extract subtle high-framerate visual features with strong audio-visual temporal cues, we rely on Segment AVCLIP \cite{synchformer2024iashin}. 
This enables our model to capture more immediate visual events with a visual framerate of 25 FPS, which is more than 6 times higher than state-of-the-art models \cite{luo2023difffoley, wang2024frierenefficientvideotoaudiogeneration}.
Following the natural co-occurrence of audio and video, we align the visual features temporally with the auditory ones, enabling our model to generate more temporally aligned audio.

\vspace{1ex}\noindent\textbf{Visual embeddings.}
Segment AVCLIP employs TimeSformer \cite{bertasius2021space} as its visual feature extractor ($M_v$).
It is pretrained contrastively with audio on a sub-clip level to extract fine-grained motion features from visual events \cite{synchformer2024iashin}.
We also experimented with action recognition models, such as S3D \cite{xie2018rethinkingspatiotemporalfeaturelearning}, and ImageNet pre-trained models, such as ResNet-50 \cite{he2015deepresiduallearningimage}.
However, these feature extractors did not yield temporally-aligned or high-fidelity results.
Our visual encoder $M_v$ transforms a visual stream $V \in \mathbb{R}^{T_v \times H \times W \times 3}$, ($T_v$ is frame count, $H$ is height, $W$ is width, and 3 is RGB color channels), into a visual feature map which is projected into visual feature embeddings $\tilde{x}_v \in \mathbb{R}^{t_v \times d_v}$ with a two-layer MLP separated by a GELU \cite{hendrycks2023gaussianerrorlinearunits} non-linearity.
Adding the trainable projection improves training efficiency by keeping the weights of the heavy feature extractor $M_v$ fixed.

\vspace{1ex}\noindent\textbf{Temporal alignment of visual embeddings.}
Our initial experiments showed that conditioning the autoregressive model by prompting the audio token sequence with visual tokens results in a poor temporal alignment, which is consistent with prior results \cite{SpecVQGAN_Iashin_2021, mei2023foleygenvisuallyguidedaudiogeneration, sheffer2023iheartruecolors, du2023conditionalgenerationaudiovideo}.
Instead, we enforce the natural co-occurrence of audio and visual cues by temporally aligning them on a token level.
The visual embeddings $\tilde{x}_v$ are duplicated temporally to match the temporal dimension of audio tokens ($t_a$), yielding a sequence $x_v \in \mathbb{R}^{t_a \times d_v}$.
If $t_a/t_v$ results in a non-integer, the video sequence is padded with learnable tokens to ensure the temporal dimensions match.
During inference, the model is conditioned only on past and present video frames, avoiding unnecessary visual features.

\subsection{Neural Audio Codec}
\label{ssec:audio_tok}
We formulate the video-to-audio generation as a next token prediction problem.
To obtain the ground truth audio tokens during training, we rely on the encoder of a state-of-the-art universal audio compression model, Descript Audio Codec (DAC) \cite{kumar2023highfidelityaudiocompressionimproved}.
During inference, we use the decoder of DAC to transform the generated audio token sequence into the waveform representation.
We use a pretrained version of DAC which was trained on speech, music, and environmental sounds.

\vspace{1ex}\noindent\textbf{Audio tokenization.}
The tokenizer ($M_a$) transforms a waveform $A \in \mathbb{R}^{T_a}$ into discrete code representations $\tilde{x}_a=M_a(A)$, where $\tilde{x}_a \in \mathbb{R}^{t_a\times N_q}$ ($N_q$ is the number of residual vector quantization (RVQ) levels and $t_a$ ($\ll T_a$) is the downsampled temporal dimension).
We use a pretrained model with $N_q = 9$ and do not update the weights during training.
Following advances in music generation \cite{copet2024simplecontrollablemusicgeneration}, we apply a delay pattern where each residual level is delayed by adding one more special learnable token at the beginning of the sequence compared to the previous level, as shown with the blocks $P$ in \cref{fig:model}.
Since we predict codes across all $N_q$ codebooks at each timestep, delaying each residual level provides the model with information about the token in the preceding residual level at that same timestep.

\vspace{1ex}\noindent\textbf{Audio embeddings.}
Audio tokens from all $N_q$ RVQ-levels ($\tilde{x}_a$) are embedded with level-specific learned embedding tables ($E_i$) and summed to represent the full-band composition of the original signal \cite{kumar2023highfidelityaudiocompressionimproved}: $x_a=\sum_{i=1}^{N_q}E_i(\tilde{x}_a^i)$, where $x_a \in \mathbb{R}^{t_a \times d_a}$, and $t_a, d_a$ are the temporal and latent dimensions.

\subsection{Autoregressive Generative Model}
\label{ssec:auto_model}
The proposed autoregressive generative model takes temporally aligned visual and audio embeddings, fuses them into a cross-modal embedding sequence, and predicts the next audio tokens across all the $N_q$ codebooks. 
After generating all the audio tokens, the token sequence is decoded back to a waveform representation.
During sampling, we employ Classifier-Free Guidance (CFG) \cite{ho2022classifierfreediffusionguidance} for enhanced generation quality.

\vspace{1ex}\noindent\textbf{Cross-modal feature fusion.}
Drawing on the success of induced alignment between the condition and the generated tokens in non-autoregressive models \cite{wang2024frierenefficientvideotoaudiogeneration}, we fuse the audio and visual embeddings into joint audio-visual embeddings via channel-wise concatenation: $x_{av} = \text{concat}_c(x_a, x_v) \in \mathbb{R}^{t_a \times (d_a + d_v)}$.
Also, since the audio sequence is padded while delaying audio tokens (see \cref{fig:model}) and the visual is not, visual embeddings appear one timestep earlier, enabling the model to recognize the condition before generating audio.

\vspace{1ex}\noindent\textbf{Architecture.}
V-AURA is built on top of a GPT-style decoder-only transformer \cite{vaswani2023attentionneed} with changes outlined by Llama architecture \cite{touvron2023llamaopenefficientfoundation, touvron2023llama2openfoundation}.
We train the model to autoregressively predict the $N_q$ audio tokens of the next timestep given the sequence of joint audio-visual embeddings ($x_{av}$) accumulated by that timestep.
We employ typical cross-entropy loss. 
The ground truth is obtained by tokenizing the original waveform with DAC.

\vspace{1ex}\noindent\textbf{Classifier-Free Guidance (CFG).}
CFG \cite{ho2022classifierfreediffusionguidance} was originally proposed for the score function estimates of the diffusion models but also applies to the autoregressive domain \cite{mei2023foleygenvisuallyguidedaudiogeneration, copet2024simplecontrollablemusicgeneration, sun2024autoregressivemodelbeatsdiffusion, kreuk2023audiogentextuallyguidedaudio}.
During training, the model is conditioned on real video embeddings and empty learnable embeddings 10\% of the time.
At inference, sampling is done by combining conditional and unconditional probabilities: $\gamma\log p(x_a^{i,j}|x_{av}^{1,1}, ...,x_{av}^{i,j-1})+(1-\gamma)\log p(x_a^{i,j}|x_{av_0}^{1,1}, ..., x_{av_0}^{i,j-1})$, where $x_{av_0}$ is the fused embedding with learned empty conditioning. 
The CFG scale ($\gamma$) controls diversity and prompt alignment, with lower scales increasing diversity and higher scales yielding more prompt-aligned results.
After experimenting, we selected $\gamma = 6$. 

\section{VisualSound Dataset}
\label{sec:visualsound}
Yue \textit{et al.} \cite{yue2024moremitigatingmultimodalhallucination} show that multimodal hallucinations of Large Vision-Language Models can be reduced by filtering the harmful training data, which in turn improves the generation quality as the model does not create irrelevant content.
Motivated by their finding, we aim to strengthen the relevance and temporal alignment between generated audio and conditional video by curating our training data. 

We propose VisualSound, a subset of the VGGSound \cite{chen2020vggsound} dataset, filtered for samples with high audio-visual correspondence.
The original dataset consists of $\sim$200k 10-second YouTube videos spanning over 300+ classes.
However, since the original dataset is constructed from in-the-wild videos with low audio-visual correspondence filtering, audio events in some samples exhibit low to no relevance to the events in the visual stream.
For example, the original audio can be replaced with non-related audio like background music, narration, audio effects, or polluted with background sounds. 

We rely on the ImageBind model \cite{girdhar2023imagebind} to identify videos with poor audio-visual correspondence.
ImageBind is a state-of-the-art joint embedding model that can embed audio and visual streams into the same feature space, allowing computing the similarity between the modalities using cosine distance.

We experiment with various similarity thresholds and proceed to train the model.
The proposed dataset has a threshold of 0.3, comprising 77\,265 training, 8\,357 validation, and 5\,425 test samples.
We release VisualSound on our project page.

\section{Experiments}
\label{sec:experiments}
\subsection{Datasets and Compared Methods}
\label{ssec:datasets}
We train our model on the VisualSound dataset, proposed in \cref{sec:visualsound}.
For evaluation, we use VGGSound-Sparse \cite{sparse2022iashin}, and the test sets of VGGSound, VisualSound, and Visually Aligned Sound (VAS) \cite{chen2020generating}.
In particular, the VGGSound test set enables us to evaluate the overall generation quality over a wide range of audio classes.
However, the audio-visual correspondence in the VGGSound is not guaranteed.
In contrast, VGGSound-Sparse and VisualSound give a better view of how well the model aligns sounds with actions over time, as these datasets are curated for strong audio-visual relevance. 
VGGSound-Sparse has 444 human-verified videos of visible and audible actions in 12 categories like \textit{playing tennis}, \textit{eating}, and \textit{dog barking}.
To further evaluate the capabilities of V-AURA, we run experiments on VAS following the train-test split of \cite{SpecVQGAN_Iashin_2021}.
We drop the samples from the \textit{fireworks} class since we observed that the temporal alignment with the visual actions is often missing.

The proposed model is compared against three other methods:
SpecVQGAN \cite{SpecVQGAN_Iashin_2021} represents the commonly-used autoregressive baseline, whereas Diff-Foley \cite{luo2023difffoley} and Frieren \cite{wang2024frierenefficientvideotoaudiogeneration} serve as state-of-the-art diffusion and RFM-based comparisons respectively.
All the compared methods were trained on VGGSound \cite{chen2020vggsound}, with SpecVQGAN also on VAS \cite{chen2020generating} and Diff-Foley also on partial AudioSet \cite{7952261}.
We note that Frieren is published as an ArXiv preprint.

\subsection{Implementation and Training Details}
\label{ssec:implementation}
Following \cite{sparse2022iashin}, we use H.264 and AAC video and audio encodings.
We resample the data to 25 FPS and 44100 Hz.
Video frames are scaled so that the short-side dimensionality is 256 pixels.
Our model is trained with a context length of 2.56 seconds.
During training, we pick a random sub-clip out of the original video while for evaluation and testing, we fix the sub-clip starting times to achieve comparable evaluation results across epochs and experiments.
The batch size is 16 clips per GPU and the models are trained on six NVIDIA V100 32GB GPUs for $\sim$150 epochs until convergence.
The model is optimized using AdamW-optimizer \cite{loshchilov2019decoupledweightdecayregularization} with $\beta=[0.9, 0.95]$.
Other training parameters are initialized following \cite{mckinzie2024mm1methodsanalysis}.
The training code and models will be publicly released.

\subsection{Evaluation Metrics}
\label{ssec:evaluation}
To achieve more consistent estimates, we generate 10 samples per test video (if not stated otherwise) and average predictions, similar to \cite{SpecVQGAN_Iashin_2021}.
For a fair comparison, we clip the videos from other methods to match our model’s context size of 2.56 seconds.
Following a common practice, we use the Fréchet Audio Distance (FAD) to judge overall audio quality and Kullback-Leibler Divergence (KLD) to evaluate the relevance of the ground truth audio with the generated audio.
Following \cite{mei2023foleygenvisuallyguidedaudiogeneration}, we use Image Bind (IB) to define the relevance between the conditional video stream and generated audio.

In addition, we propose a synchronization score, \textit{Sync},  as a metric of temporal alignment between the generated audio and the visual input.
To this end, we rely on a pre-trained general-purpose audio-visual synchronization model, Synchformer \cite{synchformer2024iashin}, that classifies a temporal offset between audio and visual traces into 21 classes ranging from $-2$ to $+2$ sec.\,with $0.2$-sec.\,increments.
The final metric is a mean absolute offset among all generated samples in milliseconds.

\subsection{Results}
\label{ssec:results}
\begin{figure}[t]
    \centering
    \includegraphics[width=0.45\textwidth]{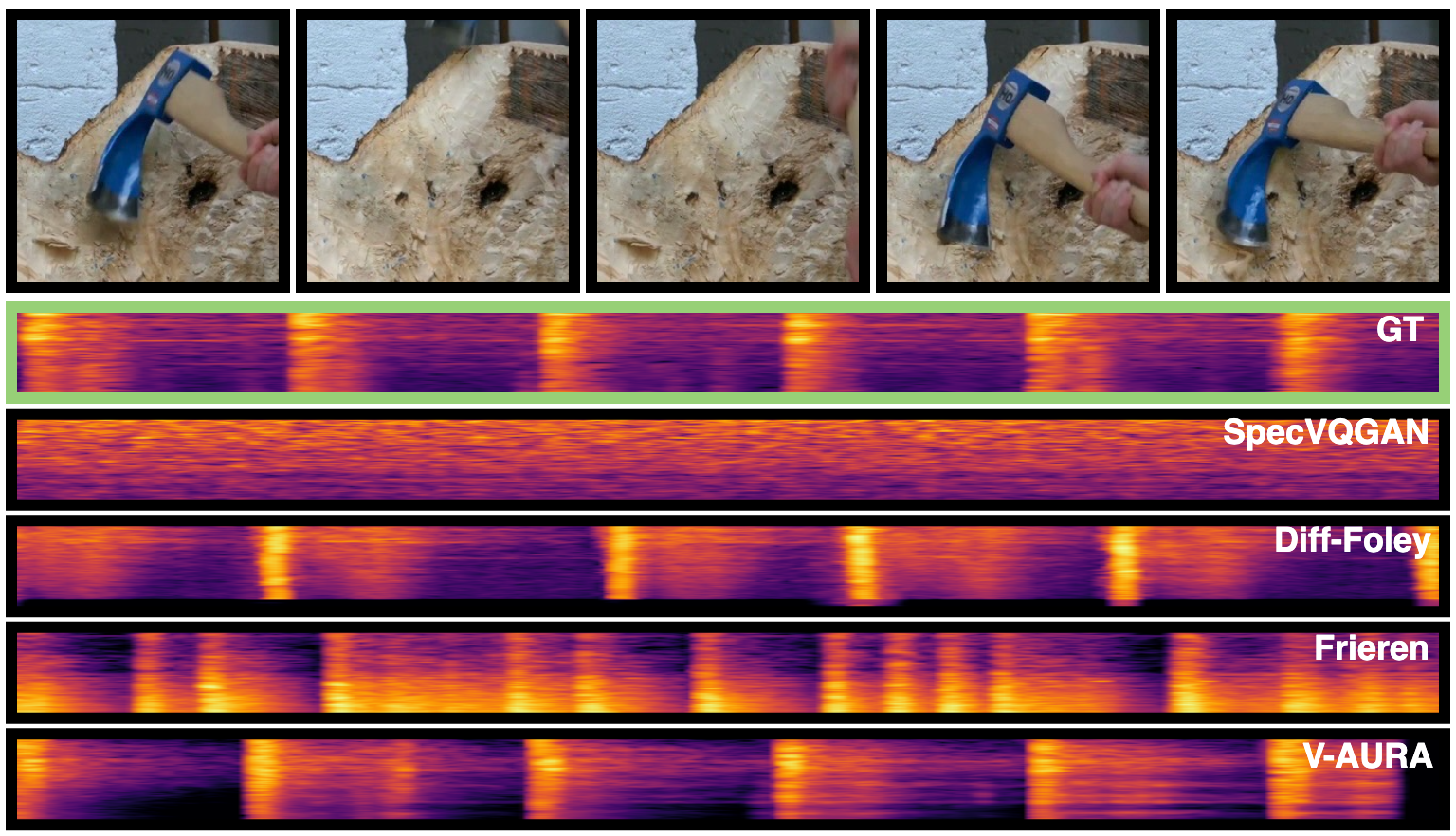}
    \vspace{-1ex}
    \caption{\textbf{V-AURA generates temporally matching audio.} Diff-Foley \cite{luo2023difffoley} misses some hits, whereas Frieren \cite{wang2024frierenefficientvideotoaudiogeneration} generates too many. SpecVQGAN \cite{SpecVQGAN_Iashin_2021} does not generate distinguishable hits.}
    \label{fig:qualitative_result}
\end{figure}

\noindent\textbf{Visually guided audio generation.}
We report the results in \cref{tab:main_res}.
We were not able to evaluate Frieren \cite{wang2024frierenefficientvideotoaudiogeneration} on VAS \cite{chen2020generating}, since the code nor samples are publicly available.
V-AURA exceeds all the compared methods in temporal quality (Sync) and relevance (KLD, IB) across all the datasets while outperforming or maintaining a comparable overall audio quality (FAD).
Especially, the evaluation on VGGSound-Sparse \cite{sparse2022iashin} highlights the V-AURA's ability to generate aligned audio (Sync) compared to other methods, whereas results on VisualSound emphasize V-AURA's strong ability to generate relevant audio (KLD, IB).
Only in terms of FAD, Fieren obtains slightly better performance. 
\Cref{fig:qualitative_result} highlights the temporal generation quality of V-AURA as all \textit{hits} are aligned with the ground truth.
Due to the low video framerate (4 FPS) of Diff-Foley \cite{luo2023difffoley} and Frieren, detecting a series of rapid hits with precise timings becomes unfeasible.
We provide more samples on our project page for subjective evaluation.

\begin{table}[t]
    \setlength{\tabcolsep}{7pt} 
    \begin{tabular}{l l rrrr}
        \multicolumn{6}{l}{\textbf{VGGSound $\downarrow$}} \\
        \toprule
        \textbf{Method} & \textbf{Type} &\textbf{KLD $\downarrow$}&\textbf{FAD $\downarrow$}&\textbf{IB $\uparrow$}&\makecell{\textbf{Sync} $\downarrow$}\\ 
        \midrule
        SpecVQGAN \cite{SpecVQGAN_Iashin_2021}                         & AR  & 3.16 & 6.41 & 10.09 & 409 \\
        Diff-Foley \cite{luo2023difffoley}                             & DM  & 3.23 & 5.62 & 16.88 & 321 \\
        Frieren \cite{wang2024frierenefficientvideotoaudiogeneration}  & RFM & 2.95 & \textbf{1.43} & 19.56 & 277 \\
        V-AURA (Ours)                                                  & AR  & \textbf{2.31} & 1.92 & \textbf{25.01} & \textbf{155} \\ 
        \bottomrule
        \multicolumn{6}{l}{\textbf{VAS $\downarrow$}} \\
        \toprule
        \textbf{Method} & \textbf{Type} &\textbf{KLD $\downarrow$}&\textbf{FAD $\downarrow$}&\textbf{IB $\uparrow$}&\makecell{\textbf{Sync} $\downarrow$}\\
        \midrule
        SpecVQGAN \cite{SpecVQGAN_Iashin_2021} & AR & 3.13 & 7.77 & 11.18 & 536 \\
        Diff-Foley \cite{luo2023difffoley}     & DM & 3.27 & 8.35 & 15.71 & 263 \\
        V-AURA (Ours)                          & AR & \textbf{1.98} & \textbf{1.98} & \textbf{29.00} & \textbf{120} \\
        \bottomrule
        \multicolumn{6}{l}{\textbf{VGGSound-Sparse $\downarrow$}} \\
        \toprule
        \textbf{Method} & \textbf{Type}&\textbf{KLD $\downarrow$}&\textbf{FAD $\downarrow$}&\textbf{IB $\uparrow$}&\makecell{\textbf{Sync} $\downarrow$}\\
        \midrule
        SpecVQGAN \cite{SpecVQGAN_Iashin_2021}                         & AR  & 3.56 & 12.93 & 11.01 & 411 \\
        Diff-Foley \cite{luo2023difffoley}                             & DM  & 2.87 & 8.92 & 22.08 & 302 \\
        Frieren \cite{wang2024frierenefficientvideotoaudiogeneration}  & RFM & 2.70 & \textbf{2.79} & 22.72 & 236 \\
        V-AURA (Ours)                                                  & AR  & \textbf{1.93} & 3.55 & \textbf{28.92} & \textbf{49} \\
        \bottomrule
        \multicolumn{6}{l}{\textbf{VisualSound $\downarrow$}} \\
        \toprule
        \textbf{Method} & \textbf{Type}&\textbf{KLD $\downarrow$}&\textbf{FAD $\downarrow$}&\textbf{IB $\uparrow$}&\makecell{\textbf{Sync} $\downarrow$}\\
        \midrule
        SpecVQGAN \cite{SpecVQGAN_Iashin_2021}                         & AR  & 3.41 & 9.02 & 10.87 & 423  \\
        Diff-Foley \cite{luo2023difffoley}                             & DM  & 2.84 & 7.24 & 19.79 & 338 \\
        Frieren \cite{wang2024frierenefficientvideotoaudiogeneration}  & RFM & 2.45 & \textbf{2.03} & 23.39 & 285 \\
        V-AURA (Ours)                                                  & AR  & \textbf{1.76} & 3.27 & \textbf{29.50} & \textbf{138} \\
        \bottomrule
    \end{tabular}
    \vspace{-1ex}
    \caption{
        \textbf{V-AURA outperforms or achieves comparable results in visually guided audio generation compared to the state-of-the-art.} Type denotes the type of the generative model: autoregressive (AR), diffusion (DM), or rectified flow matching (RFM). Results on VAS \cite{chen2020generating} were calculated over 3 samples. We could not evaluate Frieren on VAS as its code or samples are not publicly available. Out of all the models, only SpecVQGAN was trained also on VAS.
    }
    \label{tab:main_res}
\end{table}

\vspace{1ex}\noindent\textbf{Different VisualSound variants.}
\Cref{sec:visualsound} presents the novel video dataset with high audio-visual relevance, where the relevance is defined as the cosine similarity between audio and visual embeddings calculated with ImageBind \cite{girdhar2023imagebind}.
\cref{tab:ablation_simscore} compares the performance of our model on VGGSound-Sparse \cite{sparse2022iashin} after training on the dataset with increasing filtering level on cosine similarity from 0.0 to 0.4 and more.
We make the following observations:
1) The generated samples are more relevant (IB, KLD) and temporally aligned (Sync) to the visual conditioning as the noisy training samples are filtered.
2) As the dataset gets smaller, the overall quality (FAD) slightly deteriorates.
We believe that it is due to the model's inability to produce audio that would reflect the underlying probability distribution of the original unfiltered dataset.
3) The temporal alignment performance is maximised at 0.3, and filtering the dataset more makes it too small to learn meaningful representations and generalize across hundreds of data classes.
4) As the dataset size drops, the training time reduces, while the generation performance improves or remains comparable.

\begin{table}[t]
  \setlength{\tabcolsep}{4pt}
    \begin{tabular}{lrr|rrrr}
        \toprule
        \textbf{Thresh.} & \textbf{\# Samples} & \textbf{Train} \faClockO $\;\downarrow$ & \textbf{KLD $\downarrow$} & \textbf{FAD $\downarrow$} & \makecell{\textbf{IB $\uparrow$}} & \makecell{\textbf{Sync} $\downarrow$}  \\ 
        \midrule
        0.0 & 155\,591 & 708 & 1.94 & 3.16 & 28.51 & 60 \\ 
        0.2 & 119\,469 & 662 & 1.94 & 3.49 & 28.66 & 59 \\ 
        \textbf{0.3} & \textbf{77\,265} & \textbf{278} & \textbf{1.93} & \textbf{3.55} & \textbf{28.92} & \textbf{49} \\ 
        0.4 & 33\,225 & 168 & 1.97 & 4.11 & 27.31 & 71 \\ 
        \bottomrule
    \end{tabular}
    \vspace{-1ex}
    \caption{\textbf{Removing samples with low audio-visual correspondence allows us to reduce the training time and increase relevance and temporal quality.} The bolded row indicates the preferred threshold. A higher threshold indicates greater similarity between the corresponding audio and visual embeddings in the dataset. Reported samples are training samples and the training time is in GPU hours. Models were evaluated on VGGSound-Sparse \cite{sparse2022iashin}.}
    \label{tab:ablation_simscore}
\end{table}

\begin{table}[!t]
    \setlength{\tabcolsep}{10pt}
    \begin{tabular}{lrrrrrr}
        \toprule
        \textbf{Cond. methods} & \textbf{KLD $\downarrow$} & \textbf{FAD $\downarrow$} & \makecell{\textbf{IB $\uparrow$}} & \makecell{\textbf{Sync} $\downarrow$}  \\ 
        \midrule
        Prepend & 1.94 & 4.11 & 26.44 & 105 \\
        \textbf{Fusion} & \textbf{1.93} & \textbf{3.55} & \textbf{28.92} & \textbf{49} \\
        \bottomrule
    \end{tabular}
    \vspace{-1ex}
    \caption{\textbf{Cross-modal feature fusion improves synthesized audio.} The bolded row indicates the preferred conditioning method. Models were evaluated on VGGSound-Sparse \cite{sparse2022iashin}.}
    \label{tab:ablation_cond}
\end{table}

\begin{table}[!t]
    \setlength{\tabcolsep}{12pt}
    \begin{tabular}{lrrrrrr}
        \toprule
        \textbf{CFG-scale} & \textbf{KLD $\downarrow$} & \textbf{FAD $\downarrow$} & \makecell{\textbf{IB $\uparrow$}} & \makecell{\textbf{Sync} $\downarrow$}  \\ 
        \midrule
        1 & 2.48 & 11.44 & 16.94 & 155 \\
        3 & 2.04 & 5.30 & 26.71 & 80 \\
        5 & 1.94 & 3.75 & 28.52 & 52 \\
        \textbf{6} & \textbf{1.91} & \textbf{3.50} & \textbf{28.97} & \textbf{50} \\
        7 & 1.91 & 3.74 & 29.16 & 55 \\
        9 & 1.91 & 4.08 & 29.66 & 53 \\
        \bottomrule
    \end{tabular}
    \vspace{-1ex}
    \caption{\textbf{CFG-scale significantly impacts the generated audio quality.} A scale of 6 (bolded) is preferred since it produces a good balance between the metrics. Results were calculated over 3 VGGSound-Sparse \cite{sparse2022iashin} samples.}
    \label{tab:ablation_cfg}
\end{table}

\pagebreak
\subsection{Ablations}
\label{ssec:ablations}
\noindent\textbf{Conditioning methods.}
\Cref{tab:ablation_cond} shows the impact of different conditioning methods on the performance of V-AURA.
Previous autoregressive methods have prepended the conditional embeddings to the audio embedding sequence \cite{SpecVQGAN_Iashin_2021, mei2023foleygenvisuallyguidedaudiogeneration, sheffer2023iheartruecolors, du2023conditionalgenerationaudiovideo}.
We observe that fusing the modalities allows V-AURA to generate more relevant (IB, KLD), better temporally aligned (Sync), and higher quality (FAD) audio.

\vspace{1ex}\noindent\textbf{Classifier-Free Guidance scale.}
\Cref{tab:ablation_cfg} illustrates the effect of CFG-scale to the performance of V-AURA.
We emphasize temporal alignment (Sync) and thus select $\gamma = 6$.
Also, the overall generation quality (FAD) reaches the highest with the same scale.

\section{Conclusion}
\label{sec:conclusion}
We introduced V-AURA, an autoregressive video-to-audio model that generates audio that is both temporally and semantically aligned with the conditional video stream.
V-AURA consistently outperforms or achieves comparable performance with the current state-of-the-art methods, showing significant improvements in relevance and temporal accuracy.
Improvements are achieved with a high-framerate visual feature extractor combined with a cross-modal feature fusion that emphasises the natural co-occurrence of audio and visual events better than conventional conditioning methods.
Also, training V-AURA using a dataset curated for strong audio-visual correspondence mitigates hallucinations and improves the relevance and temporal quality of synthesised audio.
We proposed the dataset as a novel benchmark for video-to-audio models and refer to it as VisualSound.
Additionally, we proposed an objective temporal alignment metric for the average offset between conditional video and generated audio.

\bibliographystyle{IEEEtran}
\bibliography{refs }

\end{document}